\documentclass[letterpaper, 10 pt, conference]{ieeeconf}  % Comment this line out if you need a4paper

\IEEEoverridecommandlockouts                              % This command is only needed if 
\usepackage{graphics} % for pdf, bitmapped graphics files
\usepackage{epsfig} % for postscript graphics files
\usepackage{amssymb}  % assumes amsmath package installed

\usepackage{multirow}
\usepackage{cite}
\usepackage{booktabs}
\usepackage{tabularx}
\usepackage{xcolor}
\usepackage{siunitx} 

\newcommand{\best}[1]{\textbf{#1}}
\newcommand{\second}[1]{\underline{#1}}

\definecolor{CATBlue}{HTML}{3E5CA4}
\definecolor{CATBlueDark}{HTML}{2F4F8F}
\definecolor{CATRed}{HTML}{C5232A}

\makeatletter
\let\NAT@parse\undefined
\makeatother

\usepackage[
    colorlinks=true,
    linkcolor=CATBlueDark,
    citecolor=CATBlueDark,
    urlcolor=CATBlueDark
]{hyperref}

\usepackage[capitalise,nameinlink]{cleveref}

\crefname{figure}{Fig.}{Figs.}
\Crefname{figure}{Figure}{Figures}
\crefname{table}{Table}{Tables}
\Crefname{table}{Table}{Tables}
\crefname{equation}{Eq.}{Eqs.}
\Crefname{equation}{Eq.}{Eqs.}
\crefname{section}{Sec.}{Secs.}
\Crefname{section}{Sec.}{Secs.}

\title{\LARGE \bf
Towards Capability-Aware Traversability Navigation\\
for Unstructured Environments
}

\author{Gianluca Capezzuto, 
        Felipe Tommaselli,
        Matheus P. Angarola,
        Ricardo V. Godoy,
        and Marcelo Becker%
\thanks{This work was supported by the S\~ao Paulo Research Foundation~(FAPESP), Grants \#2025/22381-3, \#2025/20858-7, and \#2025/04308-7; and by Petr\'{o}leo Brasileiro S/A - Petrobras, using resources from the ANP R\&D clause, in partnership with the University of S\~{a}o Paulo~(USP) and the Funda\c{c}\~{a}o de Apoio \`{a} F\'{\i}sica e \`{a} Qu\'{\i}mica~(FAFQ), under Cooperation Agreements \#2023/00016-6 and \#2023/00013-7.}% 
\thanks{All authors are with the Mobile Robotics Group, University of S\~ao Paulo, S\~ao Carlos, SP, Brazil.}%
\thanks{Correspondence to \href{mailto:gianlucacapezzuto@usp.br}{\fontfamily{qcr}\selectfont gianlucacapezzuto@usp.br}}
}

\begin{document}

\maketitle
\thispagestyle{empty}
\pagestyle{empty}

%%%%%%%%%%%%%%%%%%%%%%%%%%%%%%%%%%%%%%%%%%%%%%%%%%%%%%%%%%%%%%%%%%%%%%%%%%%%%%%%
\begin{abstract}

Estimating traversability in unstructured environments requires conditioning on robot embodiment, as the same terrain can be traversable for one platform and unsafe for another. Existing methods often transfer predictions across morphologies through late-stage trajectory filtering rather than encoding platform constraints in the learned representation. We propose Capability-Aware Traversability (CAT), a framework that embeds physical limits directly into the spatial feature space. CAT grounds dense supervision masks in physical trajectories through an interactive annotation pipeline and modulates semantic terrain maps with robot-specific traversability vectors through Spatially-Adaptive Denormalization (SPADE) blocks. Across human-annotated and trajectory-aligned datasets, CAT leads all ranking-based metrics, improving AUROC by 11.0\% on physically executed trajectories and AUPRC by 15.8\% on human traces over the strongest baseline. Ablations show that spatial conditioning and per-robot prototypes produce capability sensitivity beyond generic path prediction. Deployments on a legged quadruped and a wheeled skid-steer demonstrate embodiment-aware obstacle avoidance on embedded hardware at 4.8\,Hz. Project page: \href{https://capability-aware-traversability.github.io/}{\nolinkurl{capability-aware-traversability.github.io}}

\end{abstract}

%%%%%%%%%%%%%%%%%%%%%%%%%%%%%%%%%%%%%%%%%%%%%%%%%%%%%%%%%%%%%%%%%%%%%%%%%%%%%%%%
\section{INTRODUCTION}

Semantic perception provides an essential foundation for safe navigation in the real world. Spotting a muddy trail, one intuitively categorizes visual information from the environment to avoid slipping. When navigating unfamiliar terrain, humans intuitively look beyond geometry, assessing not only the environment itself but also how specific physical capabilities interact with the surroundings~\cite{gibson1979ecological}. Similarly, robots are expected to replicate such capability awareness by mastering traversability prediction. 

Traversability estimation projects environmental features into navigation costs to determine the feasibility of traversing a specific terrain for a given robot morphology. However, merging interaction experiences across multiple platforms introduces conflicting ground-truth labels since distinct robots possess unique physical capabilities~\cite{eder2024robot}. For instance, trajectories illustrating a quadruped navigating stairs are fundamentally unsuitable for a wheeled robot. Such embodiment-specific contradictions prevent the creation of a shared traversability representation while severely limiting policy transferability across robotics platforms~\cite{gasparino2022wayfast, gasparino2024wayfaster, eder2024robot}.

Recent approaches relying solely on prior robot experience are notable, but performance often degrades outside specific training domains~\cite{gasparino2024wayfaster, wellhausen2019walk}. As Vision-Language Models (VLMs) advance, zero-shot traversability works have emerged as a natural alternative~\cite{weerakoon2025behav, gummadi2025zest}. Yet, VLMs exhibit impressive semantic reasoning that zero-shot approaches apply only to robot-agnostic scene context, so embodiment-specific feasibility never enters the learned representation. Consequently, current pipelines integrate a secondary filter into perception to discard unfeasible paths, rather than encoding embodiment in the representation itself~\cite{castro2025vamos, lee2025terrain}.

\begin{figure}[!t]
    \centering
    \includegraphics[width=\linewidth]{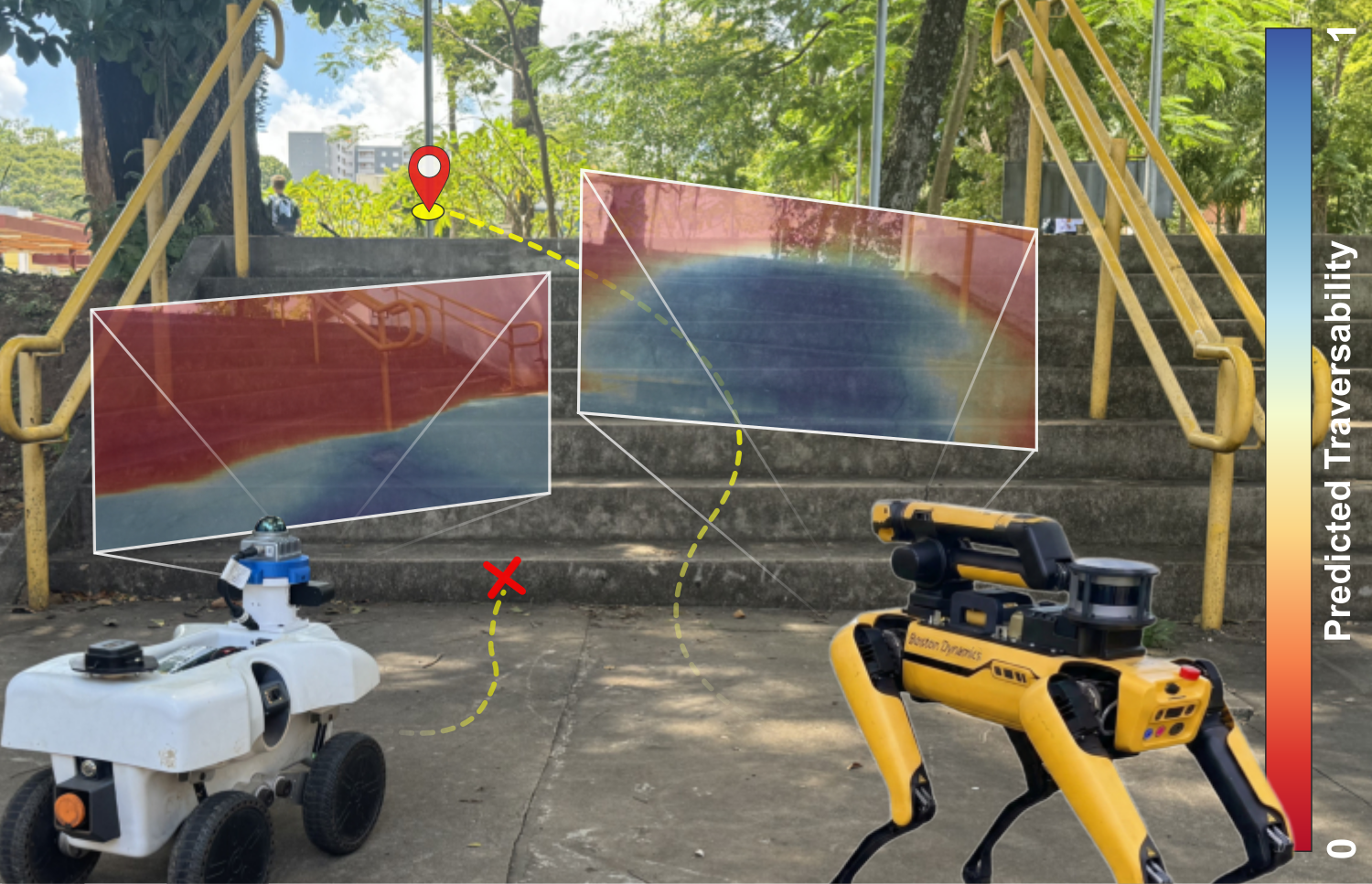}
    \caption{\textbf{Capability-Aware Traversability.} A traversability prediction framework that estimates dense navigation costs from a single observation while conditioning on robot embodiment, allowing different types of robots to assign different costs to the same terrain.}
    \label{fig:cover}
\end{figure}

Even though robotics platforms operate under different embodiment profiles, the underlying relationship between semantic terrain classes and physical interaction admits a shared structure~\cite{eder2024robot}. Rather than depending on late-stage trajectory filtering, perception models can internalize physical limits by actively modulating visual features based on robot-specific profiles, thereby integrating physical constraints directly into the spatial features to provide a unified foundation for multi-embodiment traversability estimation.

\begin{figure*}[!t]
    \centering
    \includegraphics[width=\textwidth]{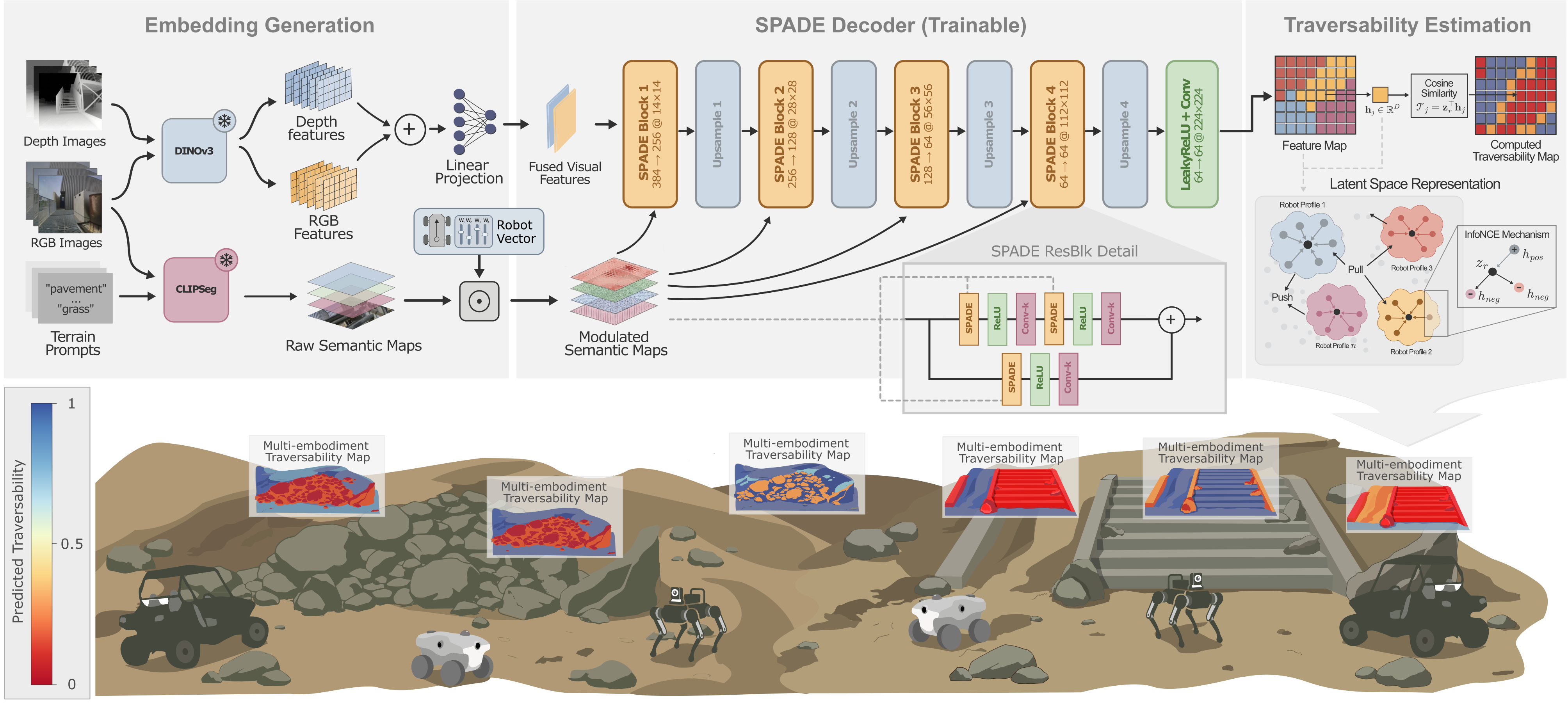}
    \caption{\textbf{Overview of Capability-Aware Traversability}. In the \textbf{Embedding Generation} stage, DINOv3~\cite{simeoni2025dinov3} encodes RGB and depth images into fused visual features, while CLIPSeg~\cite{luddecke2022clipseg} processes terrain prompts to produce raw semantic maps. To account for distinct physical constraints, a robot vector scales the raw semantic outputs to generate modulated semantic maps. The modulated semantic representation conditions a trainable \textbf{SPADE Decoder}~\cite{park2019spade} that upsamples and fuses semantics with visual features to produce a capability-conditioned spatial feature map. \textbf{Traversability Estimation} is done via similarity to robot-specific prototypes in the learned latent space, resulting in embodiment-specific traversability maps from the same scene input.}
    \label{fig:overview}
\end{figure*}

In this work, we propose \textbf{CAT} (\textbf{C}apa\-bil\-ity-\textbf{A}ware \textbf{T}ra\-ver\-sa\-bil\-ity), a unified framework for predicting traversability for distinct robot platforms in unstructured environments. To achieve this goal, we inject semantic terrain maps into our decoding process. By using Spatially-Adaptive Denormalization (SPADE) blocks~\cite{park2019spade}, CAT dynamically adapts visual embeddings to the robot's specific traversability profile~(\cref{fig:cover}). We then structure the learned latent space with a prototype per robot profile and read out traversability as the similarity between each spatial feature and its robot's prototype. Our contributions are:

\begin{itemize}
    \item A capability-conditioned feature space that embeds platform-specific limits in the perception layer, avoiding late-stage trajectory filtering;
    \item An interactive annotation pipeline that grounds zero-shot segmentation in robot trajectories, generating dense per-robot supervision masks through temporal propagation and human refinement on low-confidence frames;
    \item Improved performance on both human-aligned and trajectory-aligned protocols, with real-time obstacle avoidance on legged and wheeled robots.
\end{itemize}

\section{RELATED WORK}

\paragraph{Semantic-based Traversability} Incorporating semantic understanding into navigation stacks is a standard approach to reduce the gap between geometric sensing and environmental context~\cite{shaban2022semantic}. Recent works process semantic maps as direct input to provide contextual features to the perception module or as a distinct output layer to guide the downstream path planner~\cite{roth2024viplanner, cai2024evora, aegidius2025stepp}. Although effective for unstructured outdoor navigation, treating semantics as a direct input forces the network to learn a static mapping from visual features to traversability costs. Rather than relying on semantics as a rigid input feature, our work uses semantic representations to condition the traversability estimation. This conditioning enables the model to dynamically weigh visual cues based on the scene's semantic context.

\paragraph{Self-Supervised Traversability Learning} The robot's physical interaction with the environment has been widely used to estimate traversability. To map safe regions, previous works combine information ranging from traction~\cite{gasparino2022wayfast, gasparino2024wayfaster}, IMUs~\cite{wellhausen2019walk}, RGB cameras~\cite{mattamala2025wild, kahn2021badgr}, and LiDARs~\cite{affonso2025crow, seo2023scate}. Although robots' experience provides a highly reliable ground truth, the resulting trajectory data creates an inherently sparse supervision signal. To extract more from limited interaction data, self-supervised models correlate physically safe trajectories with dense visual representations, propagating labels from the narrow path to visually similar regions~\cite{jung2024vstrong, kim2024egocentric}. Prior methods reconstruct well-defined structures with clear trails or sidewalks, but in truly unstructured environments, purely visual similarity often correlates poorly with actual traversability. Our approach uses an interactive zero-shot process to refine propagated labels for distinct robots navigating unstructured environments.
 
\paragraph{Embodiment-Agnostic Navigation} VLMs and vision-language-action models have increasingly driven navigation frameworks toward zero-shot generalization in novel environments~\cite{weerakoon2025behav, gummadi2025zest, castro2025vamos}. Such models typically work either as high-level scene interpreters or as end-to-end path generators~\cite{cheng2025navila}. Parallel to foundation models, weakly-supervised frameworks estimate relative traversability from sparse pairwise human annotations~\cite{schreiber2024wrizz}. However, both paradigms struggle to directly internalize the mobility realities of specific robotic embodiments. Using human judgment introduces a bias, as human perception may not reflect the embodiment limits of specific platforms. Integrating VLMs into real-time navigation also remains a challenge, since inference latency and high computational cost often prove impractical for time-critical tasks such as collision avoidance. To incorporate robot-specific mobility characteristics, current architectures rely on human-ranked labels to assess terrain~\cite{schreiber2024wrizz}, or a separate filtering step to evaluate generated paths~\cite{castro2025vamos, lee2025terrain}. In contrast, our approach couples semantic context with the perception layer, internalizing physical constraints within the model.

\section{METHODS}

CAT, illustrated in \cref{fig:overview}, is an embodiment-conditioned traversability estimation framework for unstructured environments. Given an RGB-D observation, CAT predicts a dense traversability map for the target robot. To condition the prediction on robot-specific constraints, the architecture processes frozen visual embeddings alongside semantic terrain probabilities from CLIPSeg~\cite{luddecke2022clipseg} using a SPADE~\cite{park2019spade} decoder. The following subsections detail the core inference pipeline, the capability-conditioning mechanism, and the interactive annotation recipe used for training.

\subsection{Generating Traversability Labels}

To build a training set across diverse robotic navigation datasets, CAT uses the robot's past physical trajectories as the primary supervision signal. The ground-truth poses from the available state estimation are used to project the robot's trajectories into the image space of the aligned RGB-D camera, accounting for both the robot's footprint and the sensor mounting configuration. Let the 3D trajectory over a horizon $N$ be defined as the sequence $\mathcal{P} = \{P_i\}_{i=1}^N$, where each point $P_i \in \mathbb{R}^3$ represents a position in the world frame at time step $i$. Assuming homogeneous coordinates, each 3D point projects to a corresponding 2D pixel coordinate $p_i \in \mathbb{R}^2$, as described in \cref{eq:path_projection}.
\begin{equation}
    p_i = K\,T\,P_i,
    \label{eq:path_projection}
\end{equation}
where $K$ is the camera intrinsic matrix and $T$ the extrinsic transformation from the world frame to the camera frame. Projecting the full sequence $\mathcal{P}$ and filtering out occluded points by depth yields sparse positive trajectory pixels.

However, relying solely on sparse annotations limits the environment's spatial understanding. To densify the supervision, the trajectory labels are expanded into dense positive masks. Directly mapping off-the-shelf semantic segmentation to binary traversability labels is often insufficient, because a generic semantic class such as ``grass'' or ``dirt'' does not strictly correlate with safe passage, as physical traversability depends on the specific robot's capabilities and local geometric variations. Therefore, instead of treating semantic classes as rigid ground truth, our annotation pipeline generates labels through an interactive process that pairs zero-shot mask proposals with human verification on low-confidence frames.

\begin{figure}[t]
    \centering
    \includegraphics[width=\linewidth]{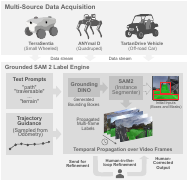}
    \caption{\textbf{Overview of the Trajectory-Grounded Label Generation Pipeline.} Data streams from the WayFASTER~\cite{gasparino2024wayfaster}, GrandTour~\cite{frey2026grandtour}, and TartanDrive~\cite{sivaprakasam2024tartandrive} datasets are processed through an interactive label engine, combining textual bounding boxes from GroundingDINO~\cite{liu2024groundingdino} with physical trajectory guidance to condition SAM 2~\cite{ravi2024sam2}. The pipeline temporally propagates the resulting dense masks across video frames and routes low-confidence segmentations to a human-in-the-loop.}
    \label{fig:pipeline_diagram}
    \vspace{-0.25in}
\end{figure}

As summarized in \cref{fig:pipeline_diagram}, given an initial RGB image, GroundingDINO~\cite{liu2024groundingdino} extracts bounding boxes that define potential traversable areas using textual prompts. The system jointly leverages the physically driven trajectory by sampling positive point prompts from the robot's projected traversal area. Both the textual bounding boxes and the physical trajectory points condition Segment Anything Model 2 (SAM 2)~\cite{ravi2024sam2}, grounding the visual segmentation in the robot's demonstrated traversals~\cite{ren2024groundedsam}. The SAM 2 inference output provides a dense traversability mask, which generates a large set of positive masks for training. Once the initial mask is generated, the model propagates the segmentation through video sequences. To ensure label integrity during propagation, the pipeline continuously monitors SAM 2 confidence scores and temporal consistency, measured by the intersection-over-union (IoU) between consecutive masks. Whenever either metric drops below a predefined threshold, the current image is sent to a human for refinement.

The pipeline generates general traversability masks rather than platform-specific labels. Robot-dependent supervision arises only during capability conditioning, where each platform’s traversability vector demotes impassable terrain and specializes the shared masks accordingly.

\subsection{Semantic Terrain Mapping and Traversability Costs}
\label{sec:semantic_trav_costs}

Robots traverse different terrains depending on their embodiment, causing the same scene to produce distinct traversability predictions for each platform. To capture the dependence, CAT decomposes terrain-dependent traversability into two complementary components: a robot-agnostic spatial semantic map of the terrain types present and a robot-specific vector encoding how each terrain type relates to a given platform's capabilities.

For semantic representation, a frozen CLIPSeg~\cite{luddecke2022clipseg} produces per-pixel terrain class probabilities. Given a set of $C$ terrain prompts, CLIPSeg outputs logits for each prompt-image pair. Applying a softmax through the prompts at each pixel results in a semantic map $\mathbf{S} \in [0,1]^{H \times W \times C}$, where each channel encodes the pixel-wise probability of the corresponding terrain class.

To couple the semantic map with the capabilities of a specific platform, each robot $r$ is characterized by a traversability vector in $\mathbb{R}^{C}$, which quantifies how suitable each terrain class is for that platform. Instead of manually engineering cost tables, we leverage a VLM to generate them automatically. We prompt the VLM with a set of uniformly sampled RGB images from the dataset to provide visual context, along with a natural language description of the robot's physical profile. The model outputs a traversability score for each terrain-robot pair, and the query runs once per dataset and robot without task-specific labels. The generated scores act as a coarse prior rather than ground truth because the network fits its predictions primarily to the recorded trajectories. A noisy class score, therefore, reshapes the conditioning without overriding the physical evidence from the terrain the robot actually traversed.

\subsection{Capability-Conditioned Traversability}

We now present the module that generates robot-specific dense traversability maps from RGB-D input. To achieve robot-specific outputs, the architecture merges frozen visual features with a robot-modulated semantic representation. More specifically, we employ DINOv3~\cite{simeoni2025dinov3} as the frozen feature encoder~$f$, leveraging its backbone’s capacity to capture rich patch-level semantics. The encoder $f$ maps the RGB image $\mathbf{x} \in \mathbb{R}^{H \times W \times 3}$ into a dense spatial feature map $\mathbf{F} \in \mathbb{R}^{h \times w \times d}$. To incorporate 3D geometric information, we replicate the single-channel depth image across three channels to match the expected input dimensions of the frozen backbone. Then, a separate inference step through $f$ extracts depth features $\mathbf{F}_{\mathrm{depth}} \in \mathbb{R}^{h \times w \times d}$. To fuse the modalities, we concatenate the feature maps along the channel dimension, forming the unified visual representation in~\cref{eq:conc_feat}.
\begin{equation}
    \label{eq:conc_feat}
    \mathbf{F}_{\mathrm{vis}} = \mathrm{Concat}(\mathbf{F}, \mathbf{F}_{\mathrm{depth}}) \in \mathbb{R}^{h \times w \times 2d}.
\end{equation}

As shown in~\cref{eq:feat_proj}, a learnable linear layer then projects the combined features to a fused dimension $d'$.
\begin{equation}\label{eq:feat_proj}
    \mathbf{F}_{\mathrm{fused}} = \mathrm{Linear}(\mathbf{F}_{\mathrm{vis}}) \in \mathbb{R}^{h \times w \times d'}.
\end{equation}

Finally, the resulting feature map $\mathbf{F}_{\mathrm{fused}}$ serves as the initial input to the decoder for generating traversability maps.

In parallel, the network modulates the semantic map by the robot's traversability vector through a per-channel multiplication that broadcasts across spatial locations. The operation scales each terrain channel by its suitability for the platform, suppressing impassable terrains while preserving traversable ones, and yields the modulated map $\hat{\mathbf{S}}_r$.

The visual features and the modulated map then converge in a progressive SPADE~\cite{park2019spade} decoder, where the modulated map acts as a spatially-varying semantic condition. The decoder consists of four SPADE residual blocks upsampling the feature map from $14 \!\times\! 14$ to $224 \!\times\! 224$. Inside each block, SPADE replaces standard batch normalization for the intermediate feature map $\mathbf{u}$, as defined by~\cref{eq:spade}.
\begin{equation}
    \mathrm{SPADE}(\mathbf{u},\, \hat{\mathbf{S}}_r)
        = \gamma(\hat{\mathbf{S}}_r) \odot \mathrm{BN}(\mathbf{u})
          + \beta(\hat{\mathbf{S}}_r),
    \label{eq:spade}
\end{equation}
where $\gamma$ and $\beta$ are learned convolutional functions of $\hat{\mathbf{S}}_r$, resized to match the current feature resolution. The decoder produces a $D$-dimensional feature map, $L_2$-normalized along the channel dimension. Because different robots generate distinct maps, the SPADE layers produce distinct modulations and output features from the same visual input. 

To structure the latent space produced by the decoder, during training, the architecture relies on a contrastive learning approach, a technique already proven highly effective for terrain traversability analysis~\cite{jung2024vstrong, seo2023learning}. For each robot profile, the network maintains a unit-norm traversability prototype $\mathbf{z}_r \in \mathbb{R}^{D}$, updated as an exponential moving average~(EMA) of the positive feature embeddings sampled for robot $r$~\cite{bu2025prototype}. At each spatial location, the network evaluates traversability through the cosine similarity between the local feature and the robot-specific prototype, defined as $\mathcal{T}_j = \mathbf{z}_r^{\!\top} \mathbf{u}_j$, where $\mathbf{u}_j \in \mathbb{R}^D$ is the decoder feature at pixel index $j$. The cosine similarity serves both to produce the traversability score map and to guide the training objective. 

The network minimizes a combined loss, defined in~\cref{eq:loss}, consisting of a trajectory-based term and a mask-based contrastive term.
\begin{equation}
    \mathcal{L} = (1 - \omega)\,\mathcal{L}_{\mathrm{traj}} + \omega\,\mathcal{L}_{\mathrm{mask}}.
    \label{eq:loss}
\end{equation}

Both $\mathcal{L}_{\mathrm{traj}}$ and $\mathcal{L}_{\mathrm{mask}}$ operate as pixel-level InfoNCE losses. The trajectory term samples positive features exclusively from the robot path pixels, whereas the mask term samples from the dense, trajectory-grounded mask pixels. Negatives are drawn from pixels outside the positive masks, together with pixels demoted by the refinement step. The InfoNCE objective pulls positive spatial features $\mathbf{u}_j$ closer to the prototype $\mathbf{z}_r$ while pushing negative spatial features away. To keep labels consistent with the physical constraints of the target robot, the pipeline refines the positive mask using the modulated semantic cost rather than the network's own output. When a pixel's modulated suitability falls below a fixed threshold, the pixel is demoted to a negative, so a region marked traversable by the segmentation but impassable by the robot's traversability vector does not act as a positive.

\section{EXPERIMENTAL RESULTS}

\begin{figure*}[!t]
    \centering
    \includegraphics[width=\textwidth]{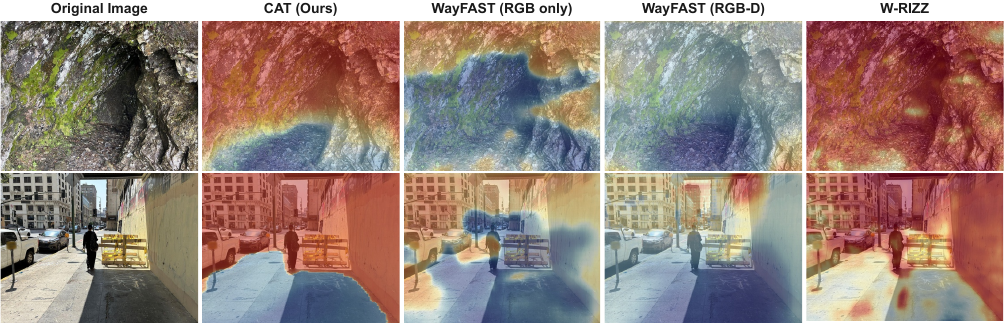}
    \caption{\textbf{Qualitative Comparison of Traversability Prediction in Unstructured Environments from NaviTrace Dataset~\cite{windecker2025navitrace}.} While baselines like WayFAST~\cite{gasparino2022wayfast} and W-RIZZ~\cite{schreiber2024wrizz} struggle to differentiate between visually similar but physically distinct terrains, CAT successfully grounds the spatial features using the semantic terrain map, resulting in cleaner separation between traversable corridors and surrounding obstacles.}
    \label{fig:qualitative_prediction}
\end{figure*}

In this section, we examine how well CAT estimates traversability for robots with different capabilities. We compare its predictions with human judgments of safe terrain and physically executed robot trajectories, and evaluate CAT on real hardware. We first describe the implementation and training setup, followed by the datasets, baselines, and metrics for quantitative evaluation. We then present the traversability results, analyze how robot capabilities affect the predictions, conduct ablation studies on the main design choices, and conclude with real-world experiments.

\subsection{Implementation Details}

CAT has 8.74\,M trainable parameters across the visual feature projection and SPADE decoder, while DINOv3 and CLIPSeg remain frozen. The decoder produces an $L_2$-normalized feature map, and traversability is computed through cosine similarity, yielding scores in $[-1,1]$ that are normalized to $[0,1]$ for visualization, with blue denoting traversable regions and red denoting obstacles.

We jointly train four robot profiles: wheeled (TerraSentia), legged (ANYmal), differential, and ATV (TartanDrive). The differential profile has no recorded trajectories and is included only through the traversability vector generated by Qwen3-VL~\cite{bai2025qwen3vl}, as described in Section~\ref{sec:semantic_trav_costs}. The InfoNCE loss is averaged across all profiles per batch, sampling 256 positive and 1024 negative pixels from the trajectory mask and 512 positive and 1024 negative pixels from the generated mask. We trained CAT for 35 epochs on a single NVIDIA L40S GPU ($\sim$8\,h), using a 70/30 training-validation split of the remaining data after holding out the test set. Training hyperparameters are listed in~\cref{tab:training}.

\subsection{Evaluation Setup}

We evaluated CAT across two distinct data distributions. First, NaviTrace~\cite{windecker2025navitrace}, a high-quality human-annotated dataset, measures whether high-confidence predictions align with human-validated safe regions. We adapt the benchmark's official trace score to CAT's dense output by converting the annotated traces into positive masks and evaluating the predicted traversability map directly against them, without requiring a downstream planner. Second, to overcome the limitations of a potentially human-biased dataset, we evaluated on a held-out 20\% test split composed of complete image sequences excluded from training. Here, positive masks are derived from the trajectories physically executed by robots and measure whether CAT assigns high traversability to the terrain that was successfully traversed. For both datasets, the target paths are expanded with a safety buffer and clipped at the image boundaries.

\begin{table}[t!]
    \caption{CAT Training Hyperparameters}
    \label{tab:training}
    \centering
    \renewcommand{\arraystretch}{1.15}
    \begin{tabularx}{\columnwidth}{@{}l>{\raggedleft\arraybackslash}X|l>{\raggedleft\arraybackslash}X@{}}
    \toprule
    \textbf{Parameter} & \textbf{Value} & \textbf{Parameter} & \textbf{Value} \\
    \midrule
    Encoder Dim.\ $d$ & 768   & Projected Dim.\ $d'$ & 384                \\
    Feature Dim.\ $D$ & 64    & Terrain Classes $C$  & 10                 \\
    Optimizer         & AdamW & Learning Rate        & $3{\times}10^{-4}$ \\
    Batch Size        & 128   & Weight Decay         & $1{\times}10^{-2}$ \\
    InfoNCE $\tau$    & 0.07  & Loss Weight $\omega$ & 0.05               \\
    EMA Momentum      & 0.99  & Trav.\ Threshold     & 0.4                \\
    \bottomrule
    \end{tabularx}
\end{table}

We benchmark CAT against three configurations from two state-of-the-art methods: WayFAST with RGB and RGB-D input~\cite{gasparino2022wayfast}, and W-RIZZ~\cite{schreiber2024wrizz}. We evaluate all baselines using their released checkpoints under the same evaluation protocol. Both checkpoints were trained on the WayFAST dataset, which contains forest-like and semi-urban outdoor scenes that broadly match the unstructured outdoor settings considered in our experiments. For both evaluations in~\cref{tab:general_prediction}, CAT uses the wheeled profile, matching the TerraSentia platform used by WayFAST and W-RIZZ. On NaviTrace, all methods are evaluated against the human traces collected for the corresponding wheeled robot.

\begin{table*}[!t]
    \centering
    \caption{Quantitative comparison of traversability prediction on NaviTrace and our held-out trajectory split}
    \label{tab:general_prediction}
    \renewcommand{\arraystretch}{1.15}
    \setlength{\tabcolsep}{4pt}
    \begin{tabular}{@{}lcccc@{\hspace{0.8em}}cccc@{}}
        \toprule
        \multirow{2}{*}{\textbf{Method}}
        & \multicolumn{4}{c}{\textbf{NaviTrace (Human-Trace Alignment)}}
        & \multicolumn{4}{c}{\textbf{Held-Out Split (Trajectory Alignment)}} \\
        \cmidrule(lr){2-5} \cmidrule(lr){6-9}
        & {\textbf{Mean Trav. $\uparrow$}}
        & {\textbf{AUROC $\uparrow$}}
        & {\textbf{AUPRC $\uparrow$}}
        & {\textbf{Ext.\,Act.@90R $\downarrow$}}
        & {\textbf{Mean Trav. $\uparrow$}}
        & {\textbf{AUROC $\uparrow$}}
        & {\textbf{AUPRC $\uparrow$}}
        & {\textbf{Ext.\,Act.@90R $\downarrow$}} \\
        \midrule
        WayFAST (RGB)~\cite{gasparino2022wayfast}
        & 0.487 & 0.525 & 0.041 & 0.756
        & 0.498 & 0.623 & 0.295 & 0.679 \\

        WayFAST (RGB-D)~\cite{gasparino2022wayfast}
        & \best{0.880} & \second{0.791} & \second{0.146} & \second{0.433}
        & \second{0.785} & 0.838 & \second{0.641} & 0.372 \\

        W-RIZZ (RGB)~\cite{schreiber2024wrizz}
        & 0.592 & 0.749 & 0.094 & 0.458
        & 0.569 & \second{0.851} & 0.554 & \second{0.284} \\

        \textbf{CAT (Ours)}
        & \second{0.770} & \best{0.794} & \best{0.169} & \best{0.424}
        & \best{0.880} & \best{0.945} & \best{0.832} & \best{0.143} \\
        \bottomrule
    \end{tabular}
\end{table*}
 
We assess prediction quality using \textit{Mean Traversability}, the average score inside the safety-buffered positive masks. \textit{Path-vs-Rest AUROC} measures how well positive-mask pixels rank above the rest of the image, while \textit{Path-vs-Rest AUPRC} evaluates the same ranking under class imbalance. \textit{Exterior Activation at 90\% Recall} selects the per-frame threshold that retains 90\% of positive-mask pixels and measures the fraction of exterior pixels above the same threshold, with lower values indicating more concentrated predictions. All metrics are computed per frame from the raw score maps and then averaged across frames. Under positive-unlabeled supervision, pixels outside the mask remain unobserved rather than verified obstacles and may still be traversable. We therefore interpret the Path-vs-Rest metrics as path-alignment measures rather than safety classification.

\begin{figure}[!t]
    \centering
    \includegraphics[width=\columnwidth]{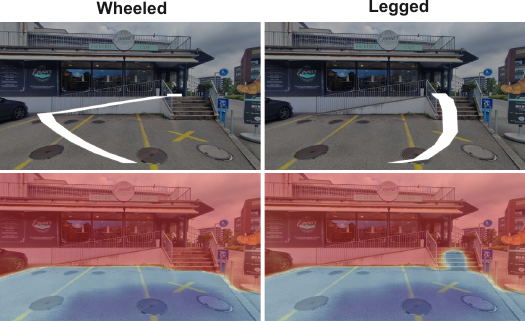}
    \caption{\textbf{Capability-Conditioned Modulation.} Qualitative traversability maps generated by CAT from the same RGB-D observation using two robot profiles. (Left) \textbf{Wheeled Profile:} CAT assigns high traversability primarily to flat terrain and avoids stairs and rough surfaces. (Right) \textbf{Legged Profile:} CAT expands the traversable region to include stairs and uneven terrain. Human annotations from NaviTrace \cite{windecker2025navitrace} are overlaid for reference, illustrating how CAT adapts its predictions to each robot's capabilities.}
    \label{fig:modulation}
    \vspace{-0.25in}
\end{figure}

To evaluate capability-specific behavior, we use NaviTrace scenes with distinct wheeled and legged annotations. For capability-conditioned modulation, we compute \textit{Mean Traversability} over each annotated path under both robot profiles. \textit{Embodiment AUROC} measures whether each profile-conditioned map ranks the region exclusive to the queried profile above the region exclusive to the other profile. Pixels shared by both annotations are excluded, and results are computed per scene and averaged across scenes.  

\subsection{Traversability Prediction}

As shown in \cref{tab:general_prediction}, CAT provides the strongest overall performance across both evaluation distributions. On our held-out trajectory split, CAT leads all metrics and improves AUROC by 11.0\% over W-RIZZ. The result shows that capability-conditioned semantic modulation generalizes to unseen sequences while separating traversed terrain from the surrounding scene. On NaviTrace, CAT achieves the highest AUROC and AUPRC together with the lowest exterior activation. The 15.8\% AUPRC improvement over WayFAST (RGB-D) shows that high responses remain more concentrated around the sparse human traces. WayFAST (RGB-D) assigns higher average scores within the trace masks, but its lower AUPRC and higher exterior activation indicate broader activation across the image. The remaining baseline results reveal different weaknesses in path discrimination and response concentration. WayFAST (RGB) poorly separates the path from surrounding surfaces. W-RIZZ ranks traversable regions more reliably than WayFAST (RGB), although its fragmented responses cover only part of the annotated path.

Qualitatively, \cref{fig:qualitative_prediction} shows that conditioning visual features with semantic terrain information resolves ambiguity in unstructured environments and produces clearer traversable corridors. In the rocky passage, CAT confines high scores to the uneven ground leading into the opening while suppressing the surrounding rock walls. WayFAST (RGB) extends high responses onto the side surfaces, WayFAST (RGB-D) activates much of the surrounding structure, and W-RIZZ recovers only isolated parts of the corridor. In the urban scene, CAT preserves a continuous corridor along the open pavement while reducing responses on the pedestrian, construction barriers, and adjacent structures. WayFAST (RGB) breaks the corridor into disconnected regions, WayFAST (RGB-D) broadens the response beyond the open pavement, and W-RIZZ highlights only a narrow sidewalk segment.

\begin{table}[!t]
    \centering
    \caption{Modulation Performance on Embodiment-Specific Scenes}
    \label{tab:modulation}
    \renewcommand{\arraystretch}{1.15}
    \begin{tabular}{@{}lcc@{}}
        \toprule
        \textbf{Profile} & \textbf{Wheeled Path Mean $\uparrow$} & \textbf{Legged Path Mean $\uparrow$} \\
        \midrule
        Wheeled Profile & 0.708 & 0.639 \\
        Legged Profile  & 0.728 & 0.730 \\
        \bottomrule
    \end{tabular}
\end{table}

\subsection{Capability-Conditioned Modulation}

Among CAT's four profiles, NaviTrace provides directly corresponding traces for the wheeled and legged embodiments, so the modulation analysis focuses on this pair. The wheeled-path evaluation includes 55 selected scenes, while the legged-path evaluation retains 73\% of the selected scenes, after excluding scenes without legged-only pixels.

As shown in~\cref{tab:modulation}, changing from the wheeled to legged profile increases the mean traversability assigned to legged paths by 14.2\%. In contrast, the same profile change produces only a 2.8\% difference on wheeled paths. The modulation effect is therefore more than four times larger on legged-specific terrain, showing that CAT responds primarily where the platform capabilities differ, as also illustrated in~\cref{fig:modulation}. The pattern is directional rather than symmetric because a legged robot can generally traverse the flat terrain available to a wheeled robot, whereas a wheeled robot cannot traverse stairs or uneven surfaces.

\subsection{Ablation Studies}

\begin{table}[!t]
    \centering
    \caption{Ablation of CAT's capability-conditioning components}
    \label{tab:ablation}
    \renewcommand{\arraystretch}{1.12}
    \setlength{\tabcolsep}{5pt}
    \begin{tabular}{@{}lccc@{}}
        \toprule
        \textbf{Variant}
        & \textbf{AUROC $\uparrow$}
        & \textbf{Emb-W $\uparrow$}
        & \textbf{Emb-L $\uparrow$} \\
        \midrule
        
        \multicolumn{4}{@{}l}{\textit{Conditioning}} \\
        \hspace{0.8em}Uniform capability
        & 0.950 & 0.543 & 0.517 \\
        \hspace{0.8em}Concatenation
        & 0.944 & 0.541 & 0.524 \\
        \hspace{0.8em}Global modulation
        & 0.951 & 0.549 & 0.530 \\
        
        \addlinespace[0.35em]
        \multicolumn{4}{@{}l}{\textit{Prototype}} \\
        \hspace{0.8em}Shared prototype
        & 0.942 & 0.540 & 0.497 \\
        \hspace{0.8em}Multiple prototypes
        & 0.938 & 0.550 & 0.521 \\
        
        \midrule
        \textbf{CAT (full)}
        & 0.945 & 0.567 & 0.540 \\
        \bottomrule
    \end{tabular}
\end{table}

In \cref{tab:ablation}, AUROC denotes Path-vs-Rest AUROC, while Emb-W and Emb-L denote Embodiment AUROC under the wheeled and legged queries, respectively. Path-vs-Rest AUROC remains similar across variants because the held-out split contains trajectories from a single wheeled platform. All configurations can therefore preserve strong path prediction even when capability sensitivity weakens. Embodiment AUROC reveals differences missed by trajectory alignment, with full CAT leading both queries.

Replacing the traversability vector with a uniform input weakens capability-specific discrimination while preserving Path-vs-Rest alignment. Concatenation and global modulation follow the same general pattern, indicating that the conditioning mechanism primarily affects profile sensitivity rather than generic path prediction. Sharing a single prototype causes the largest degradation, reducing legged Embodiment AUROC by 8.0\% relative to full CAT. Expanding the readout to three prototypes per robot does not recover the capability signal, reducing legged Embodiment AUROC by 3.5\%. The results indicate that spatial conditioning and profile-specific prototypes primarily contribute to capability awareness rather than generic path prediction.

\subsection{Real-World Deployments}

We evaluated CAT on two physical platforms: Boston Dynamics' Spot using its main RGB-D camera, and the TerraSentia skid-steer using a ZED 2i. Each deployment was designed around the capabilities of the corresponding robot, with platform-specific routes, tasks, and environments. For Spot, a trial was successful when the robot completed the forested route without colliding with surrounding obstacles. For TerraSentia, success required the robot to avoid the staircase without attempting to climb the stairs.

CAT ran at 4.8\,Hz on a Jetson Orin Nano. The predicted traversability maps were paired with a greedy pure-pursuit planner that steered toward the furthest traversable point in view, keeping the evaluation focused on the predicted maps rather than planning complexity. The frozen backbones dominated runtime, with the DINOv3 RGB and depth passes requiring 96.9\,ms and the asynchronous CLIPSeg module 53.4\,ms, versus 18.8\,ms for the trainable projection and SPADE decoder. The reported rate includes acquisition, preprocessing, synchronization, and output overhead.

\begin{figure}[!t]
    \centering
    \includegraphics[width=1.0\columnwidth]{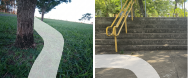}
    \caption{\textbf{Real-World Deployment of CAT.} Trials on two physical platforms, with the executed robot trajectory projected onto the onboard image frame. (Left) \textbf{Legged Profile:} Spot follows a collision-free route around the tree in a forested area. (Right) \textbf{Wheeled Profile:} TerraSentia turns away from the staircase and avoids attempting to climb the steps.}
    \label{fig:real_world}
\end{figure}

\Cref{fig:real_world} highlights two primary test cases of embodiment-aware obstacle avoidance. In the left scenario, the Spot robot accurately segmented a tree as non-traversable, generating a safe bypass trajectory through a forested area. In the right scenario, CAT correctly applied wheeled geometric constraints to mask the abrupt elevation drop of a staircase as a severe hazard, successfully forcing the TerraSentia to calculate an evasive route rather than attempting an unfeasible traversal. \cref{tab:real_world_success} reports results for both deployments. All recorded wheeled failures were due to the greedy behavior of the pure-pursuit planner rather than perceptual errors, where the planner aggressively targeted a distant traversable waypoint, causing the robot to cut through a correctly masked hazard zone to reach the distant goal.

\begin{table}[t!]
    \centering
    \caption{Real-World Capability Demonstrations}
    \label{tab:real_world_success}
    \renewcommand{\arraystretch}{1.15}
    \begin{tabular}{@{}llc@{}}
        \toprule
        \textbf{Scenario} & \textbf{Robot (Profile)} & \textbf{Successes\,/\,Trials} \\
        \midrule
        Forested Area & Spot (Legged)         & 10\,/\,10 \\
        Staircase     & TerraSentia (Wheeled) & 7\,/\,10  \\
        \bottomrule
    \end{tabular}
\end{table}

\subsection{Discussion and Limitations}

\begin{figure}[!t]
    \centering
    \includegraphics[width=1.0\columnwidth]{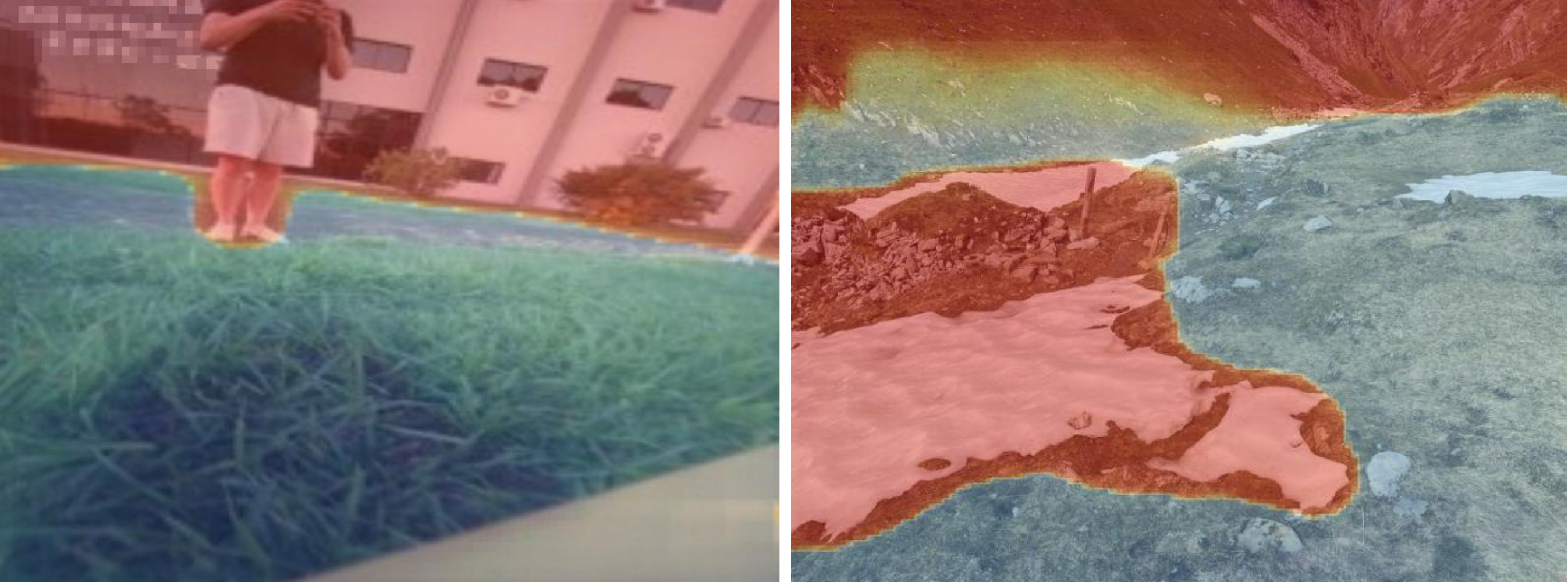}
    \caption{\textbf{CAT Findings and Limitations.} Examples of zero-shot recognition and semantic-map errors. (Left) \textbf{Zero-Shot Recognition:} CAT assigns low traversability to a person despite the absence of a corresponding training class. (Right) \textbf{Semantic Grouping Failure:} Incorrect grouping merges unrelated regions and reduces spatial detail in the prediction.}
    \label{fig:failure_case}
    \vspace{-0.25in}
\end{figure}

CAT assigns low traversability to unseen dynamic obstacles such as humans, suggesting zero-shot transfer. However, performance remains tied to the semantic map and fixed terrain prompts. Unrepresented terrain inherits the nearest class score, while grouping errors can merge unrelated regions or remove spatial detail (\cref{fig:failure_case}). Capability evaluation covers only the wheeled and legged profiles, since the differential profile lacks trajectories and NaviTrace provides neither ATV annotations nor wheeled size distinctions. Additional prototypes per robot also fail to improve capability discrimination, though the ablation does not verify that they remain distinct. Future work will collect profile-specific supervision for additional robots, characterize the latent space before revisiting multi-prototype readouts~\cite{seo2023learning}, and jointly optimize segmentation and traversability.

\section{CONCLUSION}

We proposed CAT, which reformulates traversability estimation as capability-conditioned representation learning and embeds platform constraints in spatial features rather than applying late-stage trajectory filtering. Across both protocols, CAT achieves the strongest ranking-based performance, improving AUROC by 11.0\% on physically executed trajectories and AUPRC by 15.8\% on human traces over the strongest baselines, while switching to the legged profile increases mean traversability on legged-specific paths by 14.2\%, versus 2.8\% on wheeled paths. Ablations show that spatial conditioning and per-robot prototypes drive capability sensitivity without affecting generic path prediction. We position embodiment-aware feature conditioning as a practical route toward shared cross-robot traversability.


%%%%%%%%%%%%%%%%%%%%%%%%%%%%%%%%%%%%%%%%%%%%%%%%%%%%%%%%%%%%%%%%%%%%%%%%%%%%%%%%

\bibliographystyle{IEEEtran}
\bibliography{references}

\end{document}